\documentclass{article}
\usepackage{spconf,amsmath,epsfig}
\usepackage{cite}
\usepackage{amsmath,amssymb,amsfonts}
\usepackage{algorithmic}
\usepackage{graphicx}
\usepackage{textcomp}
\usepackage{xcolor}
\usepackage{rotating}
\usepackage{multirow}
\usepackage{booktabs}
\usepackage{array}
\usepackage[colorlinks,
            linkcolor=blue,
            anchorcolor=blue,
            citecolor=blue]{hyperref}
\let\OLDthebibliography\thebibliography
\renewcommand\thebibliography[1]{
  \OLDthebibliography{#1}
  \setlength{\parskip}{0pt}
  \setlength{\itemsep}{0pt plus 0.3ex}
}

\begin{document}\sloppy

\def\x{{\mathbf x}}
\def\L{{\cal L}}

\title{BTS-NET: BI-DIRECTIONAL TRANSFER-AND-SELECTION NETWORK\\ FOR RGB-D SALIENT OBJECT DETECTION}
%
\name{Wenbo Zhang,~~Yao Jiang,~~Keren Fu\sthanks{Corresponding author (email: fkrsuper@scu.edu.cn).},~~Qijun Zhao}
\address{College of Computer Science, Sichuan University\\
   National Key Laboratory of Fundamental Science on Synthetic Vision, Sichuan University}

\maketitle

\begin{abstract}
Depth information has been proved beneficial in RGB-D salient object detection (SOD). However, depth maps obtained often suffer from low quality and inaccuracy. Most existing RGB-D SOD models have no cross-modal interactions or only have unidirectional interactions from depth to RGB in their encoder stages, which may lead to inaccurate encoder features when facing low quality depth. To address this limitation, we propose to conduct progressive bi-directional interactions as early in the encoder stage, yielding a novel bi-directional transfer-and-selection network named  BTS-Net, which adopts a set of bi-directional transfer-and-selection (BTS) modules to purify features during encoding. Based on the resulting robust encoder features, we also design an effective light-weight group decoder to achieve accurate final saliency prediction. Comprehensive experiments on six widely used datasets demonstrate that BTS-Net surpasses 16 latest state-of-the-art approaches in terms of four key metrics. 

\end{abstract}
\vspace{-0.3cm}
\section{INTRODUCTION}
\label{sec:intro}\vspace{-0.2cm}
Salient object detection (SOD) aims to locate image regions that attract much human visual attention. It is useful in many computer vision tasks, \emph{e.g.,} object segmentation \cite{SaliencyAwareVO}, tracking \cite{2019Non}, image/video compression \cite{2010A}. Though RGB SOD methods have made great progresses in recent years thanks to deep learning\cite{RGBsurvey}, they still encounter problems in challenging scenarios, \emph{e.g.,} similar foreground and background, cluttered/complex background, or low-contrast environment. 

\begin{figure}
  \centering
 \centerline{\epsfig{figure=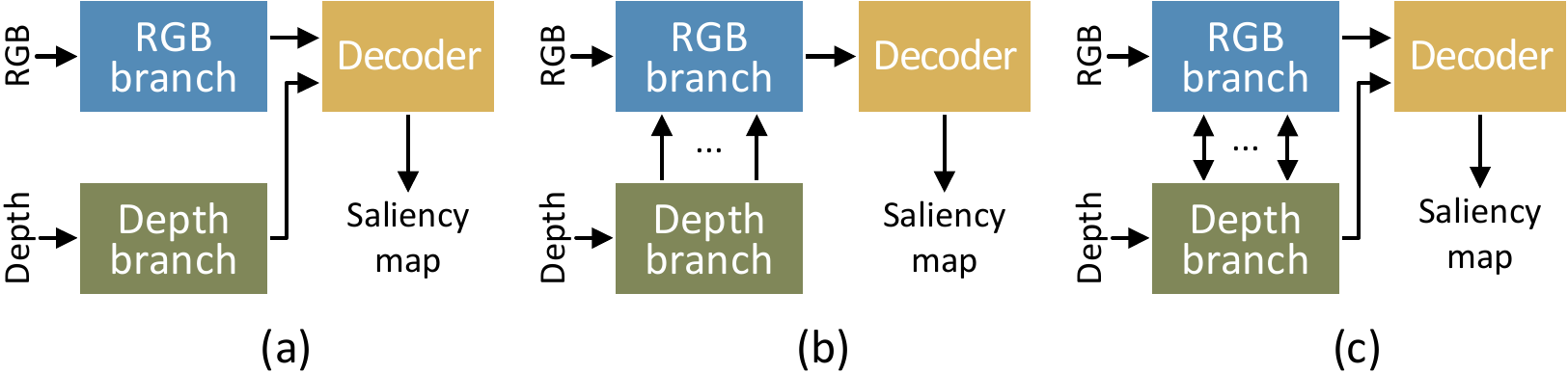,width=0.48\textwidth}}\vspace{-0.3cm}
\caption{Feature extraction strategies of existing RGB-D SOD models ((a) \cite{PCF,DRMA,JLDCF,SSF,HDFNet,PDNet} and (b)\cite{PDNet,BBSNet,CPFP}) as well as the proposed bi-directional strategy for the encoder (c).}\vspace{-0.3cm}
\label{class}
\end{figure}

With the increasing access to depth sensors, RGB-D SOD recently becomes a hot research topic \cite{BBSNet,JLDCF,UCNet,HDFNet}. Additional useful spatial information embedded in depth maps could somewhat help overcome the aforementioned challenges. 
Although a lot of advances \cite{RGBDsurvey} have been made in this field by exploring cross-modal complementarity\cite{PCF,HDFNet,UCNet,DRMA,JLDCF,SSF,cmMS,PGAR,PDNet,CPFP,BBSNet,MMCI,A2dele,CoNet,D3Net,DANet,LSSA}, we notice that existing models are still insufficient on extracting robust saliency features. As shown in Fig. \ref{class} (a) and (b), in their encoder stages, modality-aware features are usually extracted with \emph{no interactions} or \emph{unidirectional interactions}. 
For instance, in Fig.\ref{class} (a), parallel encoders\cite{PCF,DRMA,JLDCF,SSF,HDFNet,PDNet} are deployed to extract individual features of RGB and depth, and then cross-modal fusion is handled by the following decoder. In Fig. \ref{class} (b), tailor-maid sub-networks\cite{PDNet,BBSNet,CPFP} are adopted to inject depth cues into RGB as guidance/enhancement. The resulting features are then decoded to obtain the saliency map. We argue that both the above strategies may have ignored the quality issue of depth maps, since depth maps obtained from no matter depth sensors or existing datasets, are often noisy with low quality. It is obvious that in Fig. \ref{class} (a) and (b), if the input depth is inaccurate, the extracted/injected depth features will be easily affected and may degrade the final saliency map from the decoder. 

To address this issue, we propose to conduct progressive bi-directional interactions as early in the encoder, instead of late in the decoder stage. This idea is illustrated by Fig. \ref{class} (c). In this paper, we propose a novel bi-directional transfer-and-selection network, named BTS-Net, which is characterized by a new bi-directional transfer-and-selection (BTS) module applied to the encoder, enabling RGB and depth to mutually correct/refine each other as early as possible. Thus, the burden of the decoder can be well relieved. Our BTS is inspired by the attention mechanism \cite{CBAM} and cross attention \cite{CrossAtt}, and it makes features from different modalities refine each other to achieve purified features with less noise. In addition, thanks to the proposed early interaction strategy, the extracted robust hierarchical features enable us to design an effective light-weight group decoder to generate the final saliency map.

The contributions of this paper are three-fold:
\vspace{-0.2cm}
\begin{itemize}
  \item We propose BTS-Net, which is the first RGB-D SOD model to introduce bi-directional interactions across RGB and depth during the encoder stage. \vspace{-0.2cm}
  
  \item To achieve bi-directional interactions, we design a transfer-and-selection (BTS) module based on spatial-channel attention. \vspace{-0.2cm}
  
  \item We design an effective light-weight group decoder to achieve accurate final prediction.
\end{itemize}

%

\section{RELATED WORK}\vspace{-0.2cm}

The utilization of RGB-D data for SOD has been extensively explored for years. Traditional methods rely on hand-crafted features \cite{cheng2014Depth,2013An,2012Context,2015Exploiting}, while recently, deep learning-based methods have made great progress \cite{PCF,HDFNet,UCNet,DRMA,JLDCF,SSF,cmMS,PGAR,PDNet,CPFP,BBSNet,MMCI,A2dele,CoNet,D3Net,DANet}. Based on the scope of this paper, we divide existing deep-based models into two types according to how they extract RGB and depth features, namely: parallel independent encoders (Fig. \ref{class} (a)), and tailor-maid sub-networks from depth to RGB (Fig. \ref{class} (b)).

\textbf{Parallel Independent Encoders}. 
This strategy, illustrated in Fig. \ref{class} (a), first extracts features from RGB and depth images parallelly, and then fuses them using decoders. Chen \textit{et al.} \cite{PCF} proposed a cross-modal complementarity-aware fusion module. Piao \textit{et al.} \cite{DRMA} fused RGB and depth features via residual connections and refined with a depth vector and recurrent attention module. Fu \textit{et al.} \cite{JLDCF} extracted  RGB and depth features in a parallel manner but through a Siamese network. Features are then fused and refined in a densely connected manner. Zhang \textit{et al.} \cite{SSF} introduced a complimentary interaction module to select useful features. In \cite{cmMS}, Li \textit{et al.} enhanced feature representations by taking depth features as priors. Chen \textit{et al.} \cite{PGAR} proposed to extract depth features with a light-weight depth branch and conduct progressive refinement. Pang \textit{et al.} \cite{HDFNet} combined cross-modal features to generate dynamic filters, which were used to filter and enhance the decoder features. 

\textbf{Tailor-maid Sub-networks from Depth to RGB}. Recently, this unidirectional interaction from depth to RGB is introduced to the encoding stage (Fig. \ref{class} (b)), leveraging depth cues as guidance or enhancement. Zhu \textit{et al.} \cite{PDNet} used depth features extracted from a subsidiary network as a weight matrix to enhance RGB features. Zhao \textit{et al.} \cite{CPFP} computed a contrast-enhanced depth map and treated it as attention to enhance feature representations of the RGB stream. In \cite{BBSNet}, depth features were enhanced by an attention mechanism and then were fused with RGB features. Such fused features were later fed to an elaborately-designed cascaded decoder.


Different from all above methods, our BTS-Net introduces progressive bi-directional interactions in the encoder (Fig. \ref{class}(c)) to enforce mutual correction and refinement across RGB and depth branches, yielding robust encoder features.


\begin{figure*}[htb]
  \centering
 \centerline{\epsfig{figure=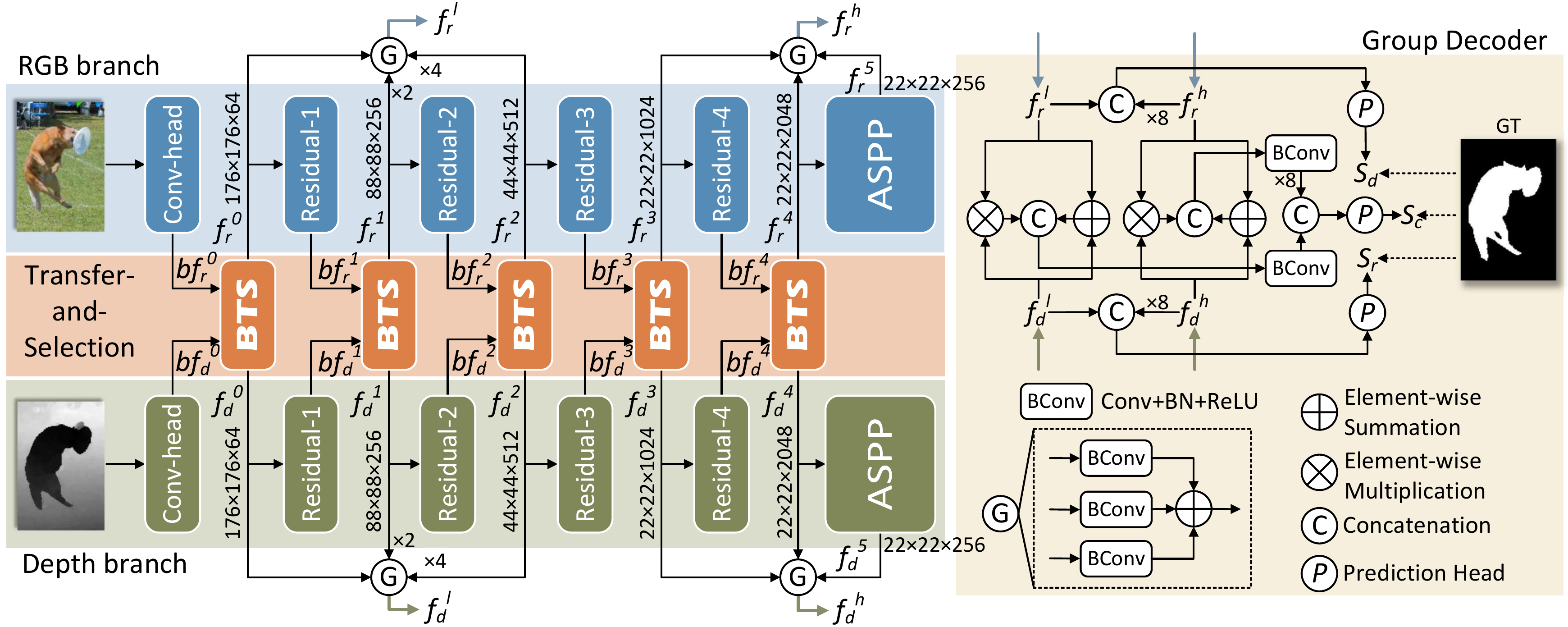,width=0.92\textwidth}}\vspace{-0.4cm}
  \caption{Block diagram of the proposed BTS-Net, which follows the typical encoder-decoder architecture. The encoder is shown on the left, whereas the decoder is shown on the right.}\vspace{-0.3cm}
  \label{blockdiagram}
\end{figure*}

\begin{figure}[htb]
  \centering
\centerline{\epsfig{figure=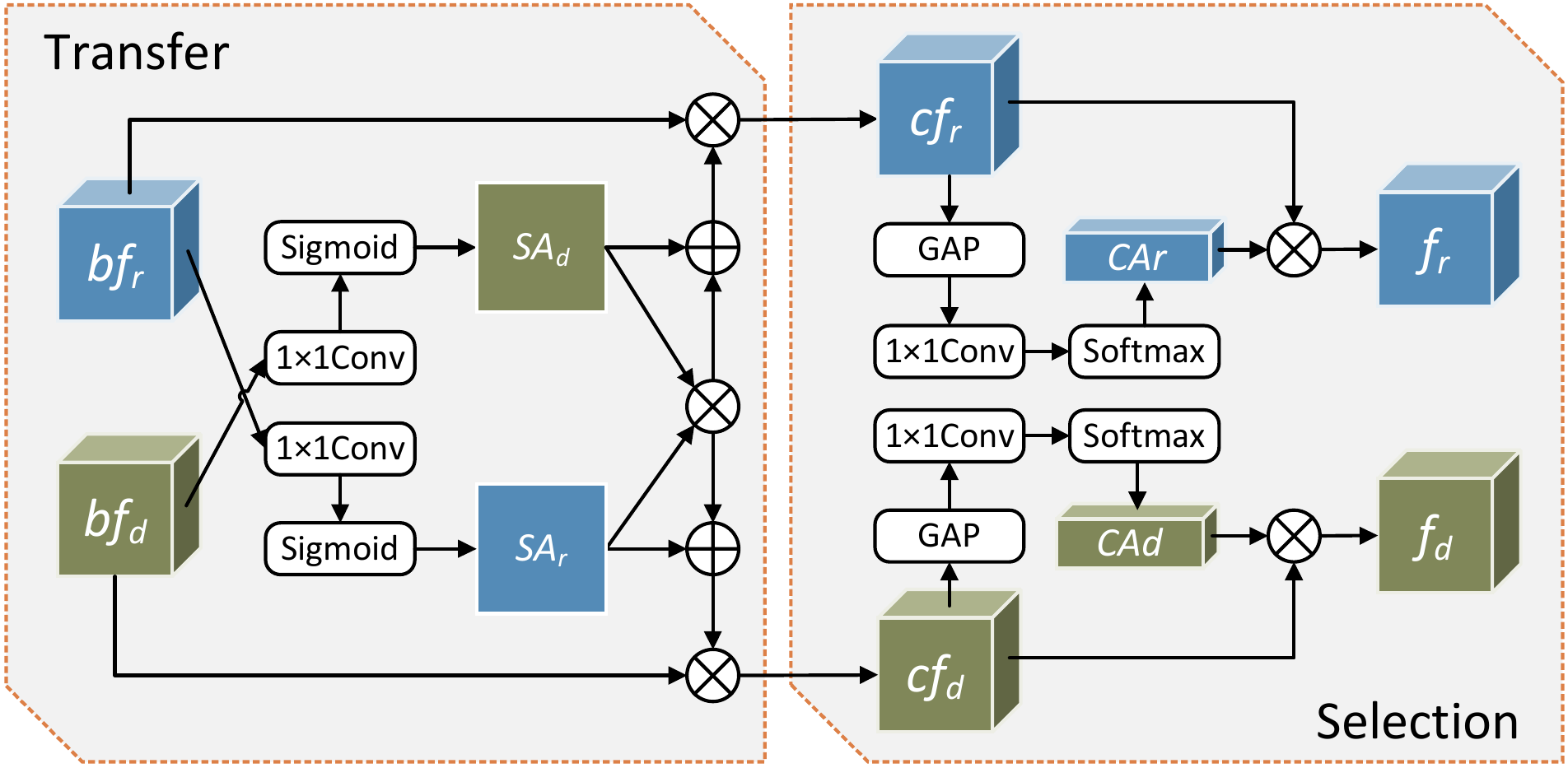,width=0.45\textwidth}}\vspace{-0.3cm}
  \caption{Detailed structure of the proposed BTS (bi-directional transfer-and-selection) module.}\vspace{-0.3cm}
  \label{BTSdiagram}
\end{figure}

\section{PROPOSED METHOD}\label{sec:method}
\vspace{-0.2cm}
Fig. \ref{blockdiagram} shows the block diagram of the proposed BTS-Net. It follows the typical encoder-decoder architecture, where the encoder is equipped with several BTS (bi-directional transfer-and-selection) modules to enforce cross-modal interactions and compensation during encoder feature extraction, resulting in hierarchical modality-aware features. Meanwhile, the decoder is elaborately designed for group-wise decoding and ultimate saliency prediction.
Specifically, the encoder consists of an RGB-related branch, a depth-related branch, and five BTS modules. These two master branches adopt the widely used ResNet-50\cite{ResNet} as backbones, leading to five feature hierarchies (note the stride of the last hierarchy is modified from 2 to 1). The input to an intermediate hierarchy in either branch is the corresponding output of the previous BTS module. Besides, in order to capture multi-scale semantic information, we add ASPP (atrous spatial pyramid pooling\cite{ASPP}) modules at the end of each branch. 

Let the feature outputs of the five RGB/depth ResNet hierarchies be denoted as $bf_{m}^{i} (m\in\{r,d\},i=0,...,4)$, and the enhanced cross-modal features from BTS modules as well as ASPPs be denoted as $f_{m}^{i} (m\in\{r,d\}, i=0,...,5)$. We regard $f_{m}^{i} (m\in\{r,d\}, i=0,1,2)$ as low-level features and $f_{m}^{i} (m\in\{r,d\}, i=3,4,5)$ as high-level features, and such low-/high-level features are then fed to the subsequent light-weight group decoder. In the followings, we describe the proposed BTS modules which can be utilized in the encoder, as well as the light-weight group decoder in detail.

\textbf{Bi-directional Transfer-and-Selection (BTS)}. The detailed structure of BTS module is shown in Fig.~\ref{BTSdiagram}, which is inspired by the well-known spatial-channel attention mechanisms\cite{CBAM}. A BTS has two processing stages: bi-directional transfer, and feature selection. Note that the former is based on spatial attention, while the latter is associated with channel attention. The underlying rational of BTS is that using spatial attention first is able to tell ``where'' a salient object is, and channel attention then can select feature channels to tell ``what'' features matter. In detail, the bi-directional transfer stage performs cross-modal attention transfer, namely applying the derived spatial attention map from either modality to the other, as illustrated in Fig.~\ref{BTSdiagram}. Since BTS is designed in a symmetric manner, for brevity, below we only elaborate the operations of transferring RGB features to the depth branch.


\begin{table*}[ht]
	\centering
	\caption{\small Quantitative RGB-D SOD results.
	$\uparrow$/$\downarrow$ denotes that a larger/smaller value is better. The best results are highlighted in \textbf{bold}.}\vspace{-0.2cm}
	\label{table:QuantitativeResults}
	\vspace{8pt}
	\footnotesize
	\renewcommand{\arraystretch}{0.7}
	\renewcommand{\tabcolsep}{0.25mm}
	\begin{tabular}{lr|cccccccccccccccc||c}
		\hline\toprule
		& \multirow{3}{*}{Metric}\centering    & PCF & MMCI & CPFP & DMRA & D3Net   &SSF &A2dele &UCNet &JL-DCF  &cmMS &CoNet &PGAR &Cas-Gnn &DANet &HDFNet &BBS-Net  & BTS-Net \\&     &\scriptsize CVPR18 &\scriptsize PR19 &\scriptsize CVPR19 &\scriptsize ICCV19 &\scriptsize TNNLS20   &\scriptsize CVPR20 &\scriptsize CVPR20  &\scriptsize CVPR20 &\scriptsize CVPR20  &\scriptsize ECCV20  &\scriptsize ECCV20 &\scriptsize ECCV20
		&\scriptsize ECCV20 &\scriptsize ECCV20 &\scriptsize ECCV20
		&\scriptsize ECCV20& \scriptsize Ours\\
		&     &\cite{PCF}&\cite{MMCI}&\cite{CPFP} &\cite{DRMA} &\cite{D3Net}   &\cite{SSF} &\cite{A2dele} &\cite{UCNet} &\cite{JLDCF} &\cite{cmMS} &\cite{CoNet} &\cite{PGAR}	&\cite{Cas-Gnn} &\cite{DANet} 
		&\cite{HDFNet} &\cite{BBSNet} &- \\
		\specialrule{0em}{1pt}{0pt}
		\hline\hline
		\specialrule{0em}{0pt}{1pt}
		\multirow{4}{*}{\begin{sideways}\textit{NJU2K}\end{sideways}}
		& $S_{\alpha}\uparrow$  & 0.877 & 0.858  & 0.879  & 0.886 & 0.900 & 0.899 &0.868	&0.897	&0.903	&0.900	&0.895	&0.909	&0.912	&0.891	&0.908	&0.921 & \bf0.921\\
		& $F_{\beta}^{\rm max}\uparrow$  & 0.872 & 0.852  & 0.877 & 0.886  & 0.950 &0.896 &0.872	&0.895	&0.903	&0.897	&0.892	&0.907	&0.916	&0.880	&0.910	&0.920 & \bf{0.924} \\
		& $E_{\xi}^{\rm max}\uparrow$     & 0.924 & 0.915  & 0.926 & 0.927  &0.950 & 0.935 &0.914	&0.936	&0.944	&0.936	&0.937	&0.940	&0.948	&0.932	&0.944	&0.949 &\bf{0.954}  \\
		& $\mathcal{M}\downarrow$  & 0.059 & 0.079  & 0.053 & 0.051 & 0.041 &0.043   &0.052	&0.043	&0.043	&0.044	&0.047	&0.042	&0.036	&0.048	&0.039	&\bf0.035 &{0.036} \\
		\midrule
		\multirow{4}{*}{\begin{sideways}\textit{NLPR}\end{sideways}}
		& $S_{\alpha}\uparrow$ & 0.874 & 0.856 &0.888 & 0.899 & 0.912 &0.914 &0.890	&0.920	&0.925	&0.915	&0.908	&0.930	&0.920	&0.915	&0.923	&0.930 &\bf{0.934} \\
		& $F_{\beta}^{\rm max}\uparrow$    & 0.841 & 0.815 &0.867 & 0.879 & 0.897 &0.896 &0.875	&0.903	&0.916	&0.896	&0.887	&0.916	&0.906	&0.901	&0.917	&0.918 &\bf{0.923} \\
		& $E_{\xi}^{\rm max}\uparrow$   & 0.925 & 0.913 &0.932 & 0.947 & 0.953 &0.953 &0.937	&0.956	&0.961	&0.949	&0.945	&0.961	&0.955	&0.953	&0.963	&0.961 &\bf {0.965} \\
		& $\mathcal{M}\downarrow$  & 0.044 & 0.059 & 0.036 &0.031	&0.025	&0.026	&0.031	&0.025	&\bf0.022	&0.027	&0.031	&0.024	&0.025	&0.029	&0.023	&0.023  &{0.023}  \\
		\midrule
		\multirow{4}{*}{\begin{sideways}\textit{STERE}\end{sideways}}
		& $S_{\alpha}\uparrow$& 0.875 & 0.873  & 0.879 & 0.835 & 0.899 &0.893  &0.885	&0.903	&0.905	&0.895	&0.908	&0.907	&0.899	&0.892 &0.900	&0.908 &\bf{0.915} \\
		& $F_{\beta}^{\rm max}\uparrow$  & 0.860 & 0.863  & 0.874 & 0.847 & 0.891 &0.890  &0.885	&0.899	&0.901	&0.891	&0.904	&0.898	&0.901	&0.881	&0.900	&0.903  &\bf{0.911} \\
		& $E_{\xi}^{\rm max}\uparrow$& 0.925 & 0.927  & 0.925 & 0.911 & 0.938 &0.936 &0.935	&0.944	&0.946	&0.937	&0.948	&0.939	&0.944	&0.930	&0.943	&0.942 &\bf{0.949} \\
		& $\mathcal{M}\downarrow$    & 0.064 & 0.068  & 0.051 & 0.066 & 0.046 &0.044  &0.043	&0.039	&0.042	&0.042	&0.040	&0.041	&0.039	&0.048	&0.042	&0.041 &\bf{0.038} \\
		\midrule
		\multirow{4}{*}{\begin{sideways}\textit{RGBD135}\end{sideways}}
		& $S_{\alpha}\uparrow$  & 0.842 & 0.848  & 0.872 & 0.900 & 0.898 &0.905&0.884	&0.934	&0.929	&0.932	&0.910	&0.913	&0.899	&0.904	&0.926	&0.933 &\bf{0.943}  \\
		& $F_{\beta}^{\rm max}\uparrow$  & 0.804 & 0.822  & 0.846 & 0.888 & 0.885 &0.883 &0.873	&0.930	&0.919	&0.922	&0.896	&0.902	&0.896	&0.894	&0.921	&0.927 &\bf{0.940}  \\
		& $E_{\xi}^{\rm max}\uparrow$  & 0.893 & 0.928  & 0.923 & 0.943 & 0.946 &0.941 &0.920	&0.976	&0.968	&0.970	&0.945	&0.945	&0.942	&0.957	&0.970	&0.966 &\bf{0.979} \\
		& $\mathcal{M}\downarrow$  & 0.049 & 0.065  & 0.038 & 0.030 & 0.031 &0.025 &0.030	&0.019	&0.022	&0.020	&0.029	&0.026	&0.026	&0.029	&0.022	&0.021 &\bf{0.018} \\
		\midrule
		\multirow{4}{*}{\begin{sideways}\textit{LFSD}\end{sideways}}
		& $S_{\alpha}\uparrow$  & 0.786 & 0.787  & 0.828 & 0.839 & 0.825 &0.859 &0.834	&0.864	&0.854	&0.849	&0.862	&0.853	&0.847	&0.845	&0.854	&0.864 &\bf{0.867}\\
		& $F_{\beta}^{\rm max}\uparrow$   & 0.775 & 0.771  & 0.826 & 0.852 & 0.810 &0.867 &0.832	&0.864	&0.862	&0.869	&0.859	&0.843	&0.847	&0.846	&0.862	&0.859 &\bf{0.874}\\
		& $E_{\xi}^{\rm max}\uparrow$   & 0.827 & 0.839  & 0.863 & 0.893 & 0.862 &0.900 &0.874	&0.905	&0.893	&0.896	&0.906	&0.890	&0.888	&0.886	&0.896	&0.901 &\bf{0.906}\\
		& $\mathcal{M}\downarrow$  & 0.119 & 0.132  & 0.088 & 0.083 & 0.095 &\bf 0.066 &0.077	&\bf0.066	&0.078	&0.074	&0.071	&0.075	&0.074	&0.083	&0.077	&0.072 &{0.070}\\
		\midrule
		
		\multirow{4}{*}{\begin{sideways}\textit{SIP}\end{sideways}}
		& $S_{\alpha}\uparrow$  & 0.842 & 0.833 &0.850 & 0.806 & 0.860 &0.874 &0.829	&0.875	&0.879	&0.867	&0.858	&0.876	&0.842	&0.878	&0.886	&0.879 &\bf{0.896} \\
		& $F_{\beta}^{\rm max}\uparrow$  & 0.838 & 0.818 &0.851 & 0.821 & 0.861 &0.880  &0.834	&0.879	&0.885	&0.871	&0.867	&0.876	&0.848	&0.884	&0.894	&0.883 &\bf{0.901} \\
		& $E_{\xi}^{\rm max}\uparrow$  & 0.901 & 0.897 &0.903 & 0.875 & 0.909 &0.921 &0.889	&0.919	&0.923	&0.907	&0.913	&0.915	& 0.890	&0.920	&0.930	&0.922 &\bf{0.933}\\
		& $\mathcal{M}\downarrow$   & 0.071 & 0.086 &0.064 & 0.085 & 0.063 &0.053 &0.070	&0.051	&0.051	&0.061	&0.063	&0.055	&0.068	&0.054	&0.048	&0.055 &\bf{0.044}\\
		\bottomrule
		\hline
	\end{tabular}
	\vspace{-8pt}
\end{table*}

Given the RGB features $bf_{r}$ at a certain hierarchy, we first compute its corresponding spatial attention map $SA_{r}$ as:
 \begin{gather}
SA_{r}=Sigmoid(Conv_{sr}(bf_{r})),
 \end{gather}
where $Sigmoid$ means the sigmoid activation function, and $Conv_{sr}$ represents a ($3\times3, 1$) convolutional layer with single-channel output.
Next, the resulting spatial attentive cue $SA_{r}$ is transferred to depth, which is mainly implemented by element-wise multiplication (denoted by mathematical symbol ``$\times$''). Before the multiplication, $SA_{r}$ is also added with a term $SA_{r} \times SA_{d}$, where $SA_{d}$ is the counterpart from the depth branch, to preserve certain modality individuality. Therefore, the depth features compensated by RGB information are formulated as: 
\begin{gather}
cf_{d}=(SA_{r}+SA_{r}\times SA_{d})\times bf_{d},
\end{gather}
where $bf_{d}$ are the corresponding depth features. After this stage, features of each modality are spatially compensated by the information from the other modality. 

Next, the obtained features $cf_{d}$ are selected along the channel dimension, which is implemented by a typical channel-attention operation\cite{CBAM}:
\begin{gather}
CA_{d}=Softmax(Conv_{cd}(GAP(cf_{d}))),\\
f_{d}=CA_{d}\times cf_{d},
\end{gather}
where $f_{d}$ indicate features that BTS outputs as in Fig.~\ref{blockdiagram} and Fig.~\ref{BTSdiagram}. $CA_{d}$ denotes the channel weight vector, $GAP$ is the global average pooling operation, $Softmax$ denotes the softmax function, and $Conv_{cd}$ is a $1\times1$ convolution, of which the input and output channel numbers are equal.
Note that after the entire transfer and selection stages, our BTS maintains the features channel and spatial dimensions. This makes the proposed BTS applicable in a ``plug-and-play'' manner in most parallel independent encoders (Fig. \ref{class} (a)). We also note that we choose not to use the widely adopted residual attention strategy \cite{ResidualAN}, which adds the attended features $f_d$ ($f_r$) with the original features $bf_d$ ($bf_r$). This is because the residual connection may limit the extent to which the complementary information can be transferred. Instead, our design allows the encoder to determine such extent adaptively. Ablation experiments in Section \ref{sec:AblationStduy} show that without this residual connection, more improvement can be obtained.   

\textbf{Group Decoder}. Our group decoder is characterized by feature grouping and three-way supervision. As well-known, deeper features from a convolutional neural network encode high-level knowledge that helps locate objects, whereas shallower features characterize low-level edge details. Our motivation of grouping is that the same-level features have better compatibility, which facilitate subsequent decoding. Therefore, after visualizing features which the encoder extracts, we roughly divide these 12 hierarchical features into four types, \emph{i.e.}, high-level RGB features ($f_{r}^{i}, i=3,4,5$), low-level RGB features ($f_{r}^{i}, i=0,1,2$), high-level depth features ($f_{d}^{i}, i=3,4,5$), and low-level depth features ($f_{d}^{i}, i=0,1,2$). 
During decoding, we first conduct feature merging within each group to save memory and computation cost. These 12 hierarchical features, denoted by $f_{m}^{i}$ with different channels, are first all
transformed to unified $k$-channel features $f_{mt}^{i}$ (in practice $k=256$) by a process consisting of convolution, BatchNorm, and ReLU. Such a process is denoted by ``BConv'' in Fig.~\ref{blockdiagram}. They are then grouped together into four types according to their properties, \emph{i.e.}, low-/high-level features and modalities, which can be defined as:\begin{gather}
f_{m}^{h}=f_{mt}^{3}+f_{mt}^{4}+f_{mt}^{5},\\
f_{m}^{l}=f_{mt}^{0}+Up(f_{mt}^{1})+Up(f_{mt}^{2}),
\end{gather}
where the subscript $m\in\{r,d\}$ indicates the RGB/depth modality, and $Up$ is the bilinear up-sampling operation. Then we utilize the grouped features $f_{m}^{h},f_{m}^{l}, m\in\{r,d\}$ to predict three ultimate saliency maps. 

To achieve fused saliency prediction $S_{c}$, we excavate cross-modal complementarity by multiplication and addition on different levels, which guarantees explicit information fusion across RGB and depth. The fused features at different levels are then concatenated and fed to a prediction head. The above operations can be summarized as:
\begin{gather}
f_{c}^{h}=BConv([f_{r}^{h}\times f_{d}^{h},f_{r}^{h}+f_{d}^{h}]),\\
f_{c}^{l}=BConv([f_{r}^{l}\times f_{d}^{l},f_{r}^{l}+f_{d}^{l}]),\\
S_{c}=P([Up(f_{c}^{h}),f_{c}^{l}]),
\end{gather}
where $P$ is a prediction head consisting of two ``BConv'', a ($1\times 1$, 1) convoluion, a Sigmoid layer, and an up-sampling operation, and $[\cdot]$ denotes the concatenation operation. $BConv$  is the ``BConv'' process mentioned before.

Moreover, in order to enhance feature learning efficacy and avoid degradation in BTS, allowing both branches to fully play their roles, we impose extra supervision to both RGB and depth branches simultaneously. The two saliency maps, namely $S_{r}$ and $S_{d}$, are generated from individual branches by using their own features:
\begin{gather}
S_{r}=P([Up(f_{r}^{h}),f_{r}^{l}]),~~S_{d}=P([Up(f_{d}^{h}),f_{d}^{l}]),
\end{gather}
where $P$, $Up$ and $[\cdot]$ are the same defined as in Eq. (7)-(9).

\textbf{Supervision}. Similar to previous works\cite{JLDCF,BBSNet,HDFNet,D3Net}, we use the standard cross-entropy loss to implement three-way supervision to $S_r$, $S_d$ and $S_c$, which is formulated as:
\begin{gather}
\mathcal{L}_{total}=\sum\limits_{m\in\{r,d,c\}}\lambda_{m}\mathcal{L}_{bce}(S_m,G)
\end{gather}
where $\mathcal{L}_{total}$ is the total loss, $\mathcal{L}_{bce}$ is the binary cross-entropy loss, $G$ denotes the ground truth, and $\lambda_{m}$ emphasizes each supervision. $\lambda_{c}=1$, $\lambda_{r}=\lambda_{d}=0.5$ are set in our experiments. During inference, $S_{c}$ is used as the final prediction result.

\section{EXPERIMENTS}
\vspace{-0.2cm}
\subsection{Datasets, Metrics and Implementation Details}\vspace{-0.2cm}
We test BTS-Net on six widely used RGB-D datasets, \emph{i.e.}, NJU2K, NLPR, STERE, RGBD135, LFSD, SIP. Following \cite{JLDCF,UCNet,CPFP}, we use the same 1500 samples from NJU2K and 700 samples from NLPR for training, and the remaining samples for testing. Four metrics are adopted for evaluation, including S-measure ($S_{\alpha}$), maximum E-measure ($E_{\xi}^{\rm max}$), maximum F-measure ($F_{\beta}^{\rm max}$), and mean absolute error (MAE, $\mathcal{M}$).
We implemented BTS-Net by Pytorch, and an input RGB-depth pair is resized to $352\times 352$ resolution. The learning rate is set to 1e-4 for the Adam optimizer, and is later degraded by 10 at 60 epochs. Batch size is set as 10, and the model is trained in total with 100 epochs. 
\vspace{-0.2cm}
\subsection{Comparison with State-of-the-Arts}\vspace{-0.2cm}
    To demonstrate the effectiveness of the proposed method, we compare it with 16 state-of-the-art (SOTA) methods, \emph{i.e.}: PCF \cite{PCF}, MMCI \cite{MMCI}, CPFP \cite{CPFP}, DMRA \cite{DRMA}, D3Net \cite{D3Net}, SSF \cite{SSF}, A2dele \cite{A2dele}, UCNet \cite{UCNet}, JLDCF \cite{JLDCF}, cmMS \cite{cmMS}, CoNet \cite{CoNet}, PGAR \cite{PGAR}, Cas-Gnn \cite{Cas-Gnn}, DANet \cite{DANet}, HDFNet \cite{HDFNet}, BBSNet \cite{BBSNet}. Quantitative results are shown in Table~\ref{table:QuantitativeResults}. It can be seen that our BTS-Net achieves superior performance over SOTAs consistently on almost all metrics.

\begin{figure}
  \centering
 \centerline{\epsfig{figure=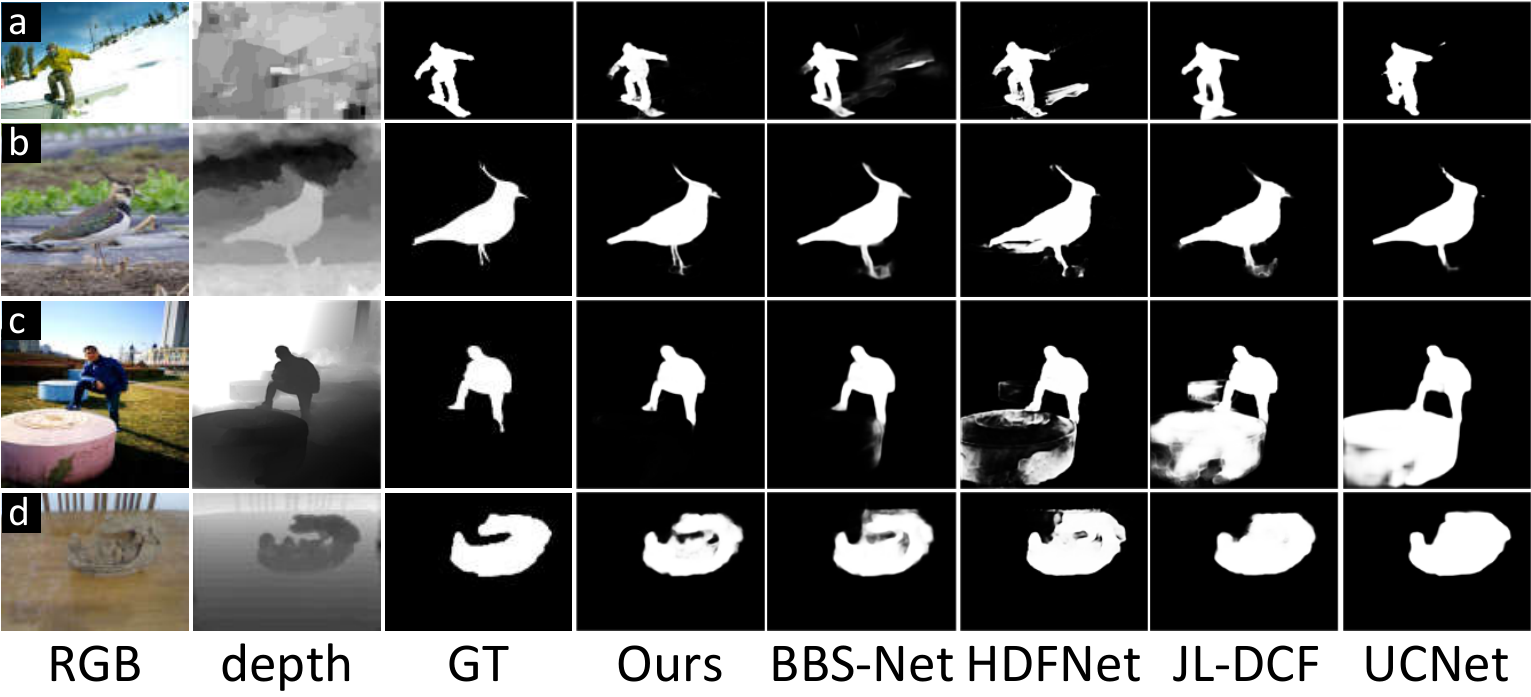,width=0.48\textwidth}} \vspace{-0.3cm}
  \caption{Visual comparisons with SOTA RGB-D SOD models.}\vspace{-0.3cm}
  \label{visualcomparison}
\end{figure}

    Fig.~\ref{visualcomparison} further shows several visual comparisons of BTS-Net with the latest representative models. From top to bottom, the quality of depth maps varies from poor to good: (a) the depth almost misses the entire object; (b) depth lacks details of the bird's head and feet; (c) depth has good contrast but non-salient regions are adjoined; (d) depth is relatively good but the RGB has low contrast. Our BTS-Net performs well and robustly in all the above cases, especially in (a) and (b), where the depth presents low quality and missing information.

\begin{table}[t]
	\centering
	\caption{Results of different interaction strategies. Details are in Section~\ref{sec:AblationStduy}: ``Interaction Directions of BTS''.}\vspace{-0.2cm}
	\label{table:AS1}
	\vspace{8pt}
	\footnotesize
	\renewcommand{\tabcolsep}{0.4mm}
	\begin{tabular}{c|c|c|ccc|ccc|ccc}
		\hline\toprule
		\multirow{2}{*}{\#} 
		&\multirow{2}{*}{Direction}
		&\multirow{2}{*}{Res}
		&\multicolumn{3}{c|}{\textbf{NJU2K}}  &\multicolumn{3}{c|}{\textbf{STERE}}
		&\multicolumn{3}{c}{\textbf{SIP}} \\ &&
		&$S_{\alpha}$
		& $F_{\beta}^{\rm max}$
		&$\mathcal{M}$ 
	    &$S_{\alpha}$
	    & $F_{\beta}^{\rm max}$
		&$\mathcal{M}$ 
		&$S_{\alpha}$
		& $F_{\beta}^{\rm max}$
		&$\mathcal{M}$ 
		
  \\
		\midrule
		1 &None &  &0.868	&0.862	&0.064	&0.730	&0.685	&0.117	&0.873	&0.875	&0.060
 \\
		2 &R$\leftarrow$D  & &0.912	&0.914	&0.040	&0.892	&0.888	&0.047	&0.891	&0.896	&0.048
  \\
		 3 &R$\rightarrow$D  & &0.920	&0.921	&\bf{0.035}	&0.911	&0.905	&0.039	&0.890	&0.895	&0.047
  \\
		4 &R$\leftrightarrow$D & &\bf{0.921}	&\bf{0.924}	&0.036	&\bf{0.915}	&\bf{0.911}	&\bf{0.038} &\bf{0.896}	&\bf{0.901}	&\bf{0.044}
\\
		 5 &R$\leftrightarrow$D &\checkmark  &0.918	&0.918	&0.037		&0.912	&0.909	&0.038	&0.890	&0.896	&0.048
  \\
		\bottomrule
		\hline
	\end{tabular}
	\vspace{-8pt}
\end{table}

\vspace{-0.5cm}
\subsection{Ablation Study}
\label{sec:AblationStduy}
\vspace{-0.1cm}

\begin{figure}
  \centering
 \centerline{\epsfig{figure=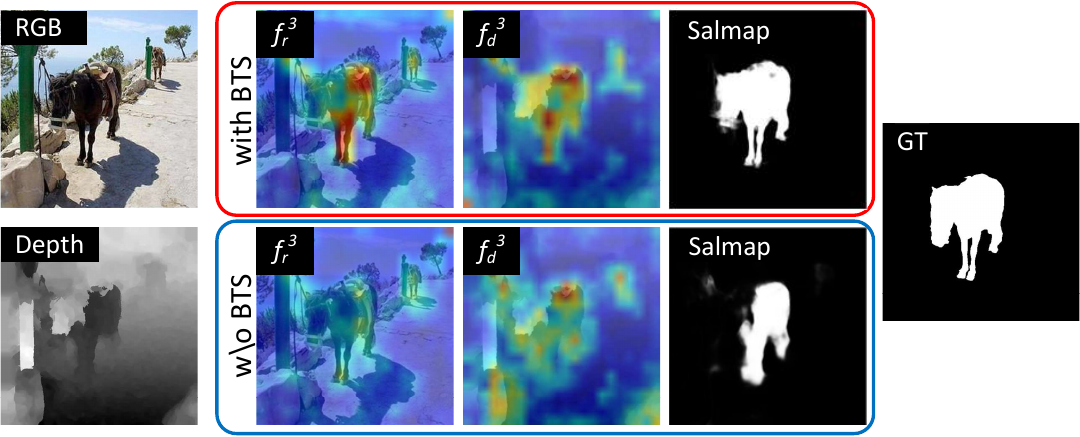,width=0.48\textwidth}} \vspace{-0.3cm}
\caption{Visualized features ($f_{r}^{3}$ and $f_{d}^{3}$ in BTS-Net) from setting \#4 (with BTS) and \#1 (w\textbackslash o BTS) in Table \ref{table:AS1}.}\vspace{-0.3cm}
\label{heatmap}
\end{figure}

\textbf{Interaction Directions of BTS.} To validate the rationality of the proposed BTS module, we set up five experiments with different settings. Notation ``R$\leftarrow$D'' means introducing depth to the RGB branch, and ``R$\rightarrow$D'' means the vice versa. ``R$\leftrightarrow$D'' means the proposed bi-directional interactions. ``Res'' means introducing residual connections into BTS for both branches as mentioned in Section \ref{sec:method}. For fair comparison, these settings are conducted by only switching connections inside BTS while keeping the main components (\emph{e.g.}, spatial and channel attention) maintained. 
Ablation results are shown in Table~\ref{table:AS1}, where row \#1 means no interaction exists between the two branches, leading to the worst results. Rows \#2 and \#3 are better than \#1, showing that uni-directional interaction is better than none. Specially, row \#3 shows much better results than \#2 on STERE dataset, indicating that transferring RGB to depth could mitigate the influence from inaccurate depth\footnote{According to our observation and also \cite{JLDCF}, the depth quality of STERE is relatively poor among the six datasets. Image (a) in Fig. \ref{visualcomparison} is from STERE.}, which rightly supports our claim. Comparing row \#4 (the default BTS) to \#2 and \#3, the improvement is consistent and notable. This validates the proposed bi-directional interaction strategy in BTS. Lastly, row \#5 leads to no boost over \#4. This may be caused by the limitation on transfer ability brought by the residual connection. Fig. \ref{heatmap} shows a comparative example between visualized features \#1 and \#4, where one can see \#4 results in more robust features as well as the final saliency map.

\begin{table}[t]
	\centering
	\caption{Results of different internal attention designs. Details are in Section~\ref{sec:AblationStduy}: ``Internal Attention Designs of BTS''.}
 	\vspace{-0.2cm}
	\label{table:AS2}
	\vspace{8pt}
	\footnotesize
	\renewcommand{\tabcolsep}{0.8mm}
	\begin{tabular}{c|ccc|ccc|ccc}
		\hline\toprule
\multirow{2}{*}{\textbf{Settings}} &\multicolumn{3}{c|}{\textbf{NJU2K}}  &\multicolumn{3}{c|}{\textbf{STERE}}
		&\multicolumn{3}{c}{\textbf{SIP}} \\
		&$S_{\alpha}$
		& $F_{\beta}^{\rm max}$
		&$\mathcal{M}$ 
	    &$S_{\alpha}$
	    & $F_{\beta}^{\rm max}$
		&$\mathcal{M}$ 
		&$S_{\alpha}$
		& $F_{\beta}^{\rm max}$
		&$\mathcal{M}$




\\
		\midrule

		 Only SA  &0.914	&0.917	&0.039	&0.903	&0.900	&0.044	&0.887	&0.892	&0.050

  \\
		 CA-SA   &0.914	&0.915	&0.039	&0.901	&0.896	&0.044		&0.892	&0.899	&0.047

  \\
		SA-CA  &\bf{0.921}	&\bf{0.924}	&\bf{0.036} &\bf{0.915}	&\bf{0.911}	&\bf{0.038}	&\bf{0.896}	&\bf{0.901}	&\bf{0.044}
  \\
		\bottomrule
		\hline
	\end{tabular}
	\vspace{-8pt}
\end{table}

\begin{table}[t]
	\centering
	\caption{Results from the U-net and our group decoder (GD). Details are in Section~\ref{sec:AblationStduy}: ``Light-weight Group Decoder''.}\vspace{-0.2cm}
	\label{table:AS3}
	\vspace{8pt}
	\footnotesize
	\renewcommand{\tabcolsep}{0.18mm}
	\begin{tabular}{c|c|ccc|ccc|ccc}
		\hline\toprule
		\multirow{2}{*}{Decoder}
		&\multirow{2}{*}{Parameters} &\multicolumn{3}{c|}{\textbf{NJU2K}}  &\multicolumn{3}{c|}{\textbf{STERE}}
		&\multicolumn{3}{c}{\textbf{SIP}} \\
		&
		&$S_{\alpha}$
		& $F_{\beta}^{\rm max}$
		&$\mathcal{M}$ 
	    &$S_{\alpha}$
	    & $F_{\beta}^{\rm max}$
		&$\mathcal{M}$ 
		&$S_{\alpha}$
		& $F_{\beta}^{\rm max}$
		&$\mathcal{M}$ 
		


\\
		\midrule
		 U-net  &32.4M&0.913	&0.913	&0.041	&0.908	&0.902	&0.042		&0.889	&0.892	&0.049
 \\
		GD-D 	&1.8M&0.912	&0.910	&0.041	&0.904	&0.895	&0.044	&0.892	&0.895	&0.047

  \\	GD-R	&1.8M&0.917	&0.918	&0.038	&0.914	&0.909	&0.039	&0.892	&0.897	&0.047

  \\	GD-C   &4.1M &\bf{0.921} &\bf{0.924}	&\bf{0.036}	 &\bf{0.915}	&\bf{0.911}	&\bf{0.038}	&\bf{0.896}	&\bf{0.901}	&\bf{0.044}
  \\
		\bottomrule
		\hline
	\end{tabular}
	\vspace{-8pt}
\end{table}

\textbf{Internal Attention Designs of BTS.} To validate the current attention design in BTS, we also set up three different experiments, whose results are shown in Table~\ref{table:AS2}. Notations ``Only SA'', ``CA-SA'', and ``SA-CA'' denote: only applying spatial attention without channel attention, changing the order of the spatial and channel attention (\emph{i.e.}, the latter comes first), and the default design of BTS-Net (\emph{i.e.}, the spatial attention comes first), respectively. Comparing the default design of BTS, namely SA-CA, to the other two variants, one can see it consistently achieves the best performance. Such comparative experiments show that the order of spatial-channel attention is crucial for introducing attention-aware interactions, whereas combining channel attention with bi-directional spatial attention transfer is effective. 

 \textbf{Light-weight Group Decoder.} To validate our light-weight group decoder, we evaluate four results. Performance during inference is shown in Table~\ref{table:AS3}, where ``U-net'' denotes the results generated by a typical U-net decoder. Basically, we concatenate RGB and depth features at the same hierarchies first and then feed the concatenated 512-channel features to a typical U-net decoder consisting of progressive upsampling, concatenation and convolution, at the end of which the same prediction head is applied to obtain the saliency map $S_c$. Note that in this experiment, the 
three-way supervision was preserved. Notation ``GD-D'', ``GD-R'', and ``GD-C'' denote the decoders deployed to obtain results $S_d$, $S_r$ and $S_c$ in BTS-Net as shown in Fig. \ref{BTSdiagram}. Note that regarding GD-D/GD-R, their parameters come mainly from the prediction heads.
 
 From Table~\ref{table:AS3}, one can see that the proposed decoder, which consists of GD-D, GD-R and GD-C, has much fewer parameters and is more light-weight. Specially, GD-C's parameters are only $\sim$12.7\% of those of the U-net, and meanwhile, GD-C which outputs $S_c$ achieves the best performance. Also note that GD-D/GD-R are even lighter, since they only involve the prediction heads. We attribute the success of the proposed group decoder partly to the efficacy of BTS modules in the encoder (with BTS, the encoder parameters increase from 80.3M to 91.5M), as the resulted robust features from the two branches make it possible for using relatively simple decoding. Also, the superior performance of GD-C comparing to GD-D and GD-R shows that fusing encoder features of the two modalities is essential for better RGB-D SOD.
 \vspace{-0.2cm}
 \section{Conclusion}\vspace{-0.2cm}
We introduce BTS-Net, the first RGB-D SOD model that adopts the idea of using bi-directional interactions between RGB and depth in the encoder. A light-weight group decoder is proposed to collaborate with the encoder in order to achieve high-quality saliency maps. Comprehensive comparisons to SOTA approaches as well as ablation experiments have validated the proposed bi-directional interaction strategy, internal designs of the BTS module, and also the group decoder. Since BTS can be applied in a ``plug-and-play'' fashion, it will be interesting to use BTS to boost existing models in the future.    

\vspace{-2pt}
\small{\vspace{.1in}\noindent\textbf{Acknowledgments.}\quad
This work was supported by the NSFC, under No. 61703077, 61773270, 61971005, the Chengdu Key Research and Development Support Program (2019-YF09-00129-GX), and SCU-Luzhou Municipal People's Government Strategic Cooperation Project (No. 2020CDLZ-10).}

\footnotesize
{
\bibliographystyle{IEEEbib}
\bibliography{icme2021template}
}
\end{document}